\documentclass{article}
\usepackage{ijcai20}

\usepackage{times}
\usepackage{soul}
\usepackage{url}
\usepackage[hidelinks]{hyperref}
\usepackage[utf8]{inputenc}
\usepackage[small]{caption}
\usepackage{graphicx}
\usepackage{array}
\usepackage{amsfonts,amssymb,amsmath,amsthm}
\usepackage{booktabs}
\usepackage{algorithm}
\usepackage{algorithmic}
\usepackage{multirow}
\usepackage{subfigure}
\newcommand{\ie}{\textit{i.e.},~}
\newcommand{\eg}{\textit{e.g.},~}

\title{Progressive Domain-Independent Feature Decomposition Network for \\Zero-Shot Sketch-Based Image Retrieval}

\author{
	Xinxun Xu$^1$\and
	Muli Yang$^1$\and
	Yanhua Yang$^1$\footnote{Contact Author}\And
	Hao Wang$^1$\\
	\affiliations
	$^1$Xidian University\\
	\emails
	xinxun.xu@gmail.com,
	muliyang.xd@gmail.com,
    yanhyang@xidian.edu.cn,
	haowang.xidian@gmail.com
}

\begin{document}
	
	\maketitle
	
	\begin{abstract}
		Zero-shot sketch-based image retrieval (ZS-SBIR) is a specific cross-modal retrieval task for searching natural images given free-hand sketches under the zero-shot scenario.
		Most existing methods solve this problem by simultaneously projecting visual features and semantic supervision into a low-dimensional common space for efficient retrieval.
		However, such low-dimensional projection destroys the completeness of semantic knowledge in original semantic space, so that it is unable to transfer useful knowledge well when learning semantic from different modalities.
		Moreover, the domain information and semantic information are entangled in visual features, which is not conducive for cross-modal matching since it will hinder the reduction of domain gap between sketch and image.
		In this paper, we propose a \underline{P}rogressive \underline{D}omain-independent \underline{F}eature \underline{D}ecomposition (PDFD) network for ZS-SBIR.
		Specifically, with the supervision of original semantic knowledge, PDFD decomposes visual features into domain features and semantic ones, and then the semantic features are projected into common space as retrieval features for ZS-SBIR.
		The progressive projection strategy maintains strong semantic supervision.
		Besides, to guarantee the retrieval features to capture clean and complete semantic information, the cross-reconstruction loss is introduced to encourage that any combinations of retrieval features and domain features can reconstruct the visual features.
		Extensive experiments demonstrate the superiority of our PDFD over state-of-the-art competitors.
		
	\end{abstract}
	
	\section{Introduction}
	With the explosive growth of image contents on the Internet, image retrieval has been playing an important role in many fields. However, conventional image retrieval requires providing textual descriptions, which are difficult to be obtained in many real-world cases. On mobile devices, image retrieval with free-hand sketches, for delivering targeted candidates visually and concisely, has attracted increasing attention and formed the area of Sketch-Based Image Retrieval (SBIR). Since it is difficult to guarantee that all categories are trained in realistic scenarios, unsatisfactory performance is often yielded when testing on unseen categories. In view of this, a more realistic setting is emerged, namely ZS-SBIR, which combines Zero-Shot Learning (ZSL) and SBIR for real-world applications.
	ZS-SBIR is extremely challenging since it simultaneously needs to deal with cross-modal matching, significant domain gap, as well as limited knowledge of unseen classes. The traditional SBIR methods~\cite{liu:deep,zhang:generative} cannot directly address these problems effectively since they over-fit the source domain and meanwhile neglect the unseen categories. On the contrary, traditional ZSL~\cite{kodirov:semantic,zhang:zero} methods often focus on solving single-modal problems. Therefore, ZS-SBIR tries to solve these problems by combining the advantages of the above methods sufficiently.
	
	\begin{figure}[!t]
		\centering
		\subfigure[]{
			\begin{minipage}[t]{2.7cm}
				\centering
				\includegraphics[width=2.7cm]{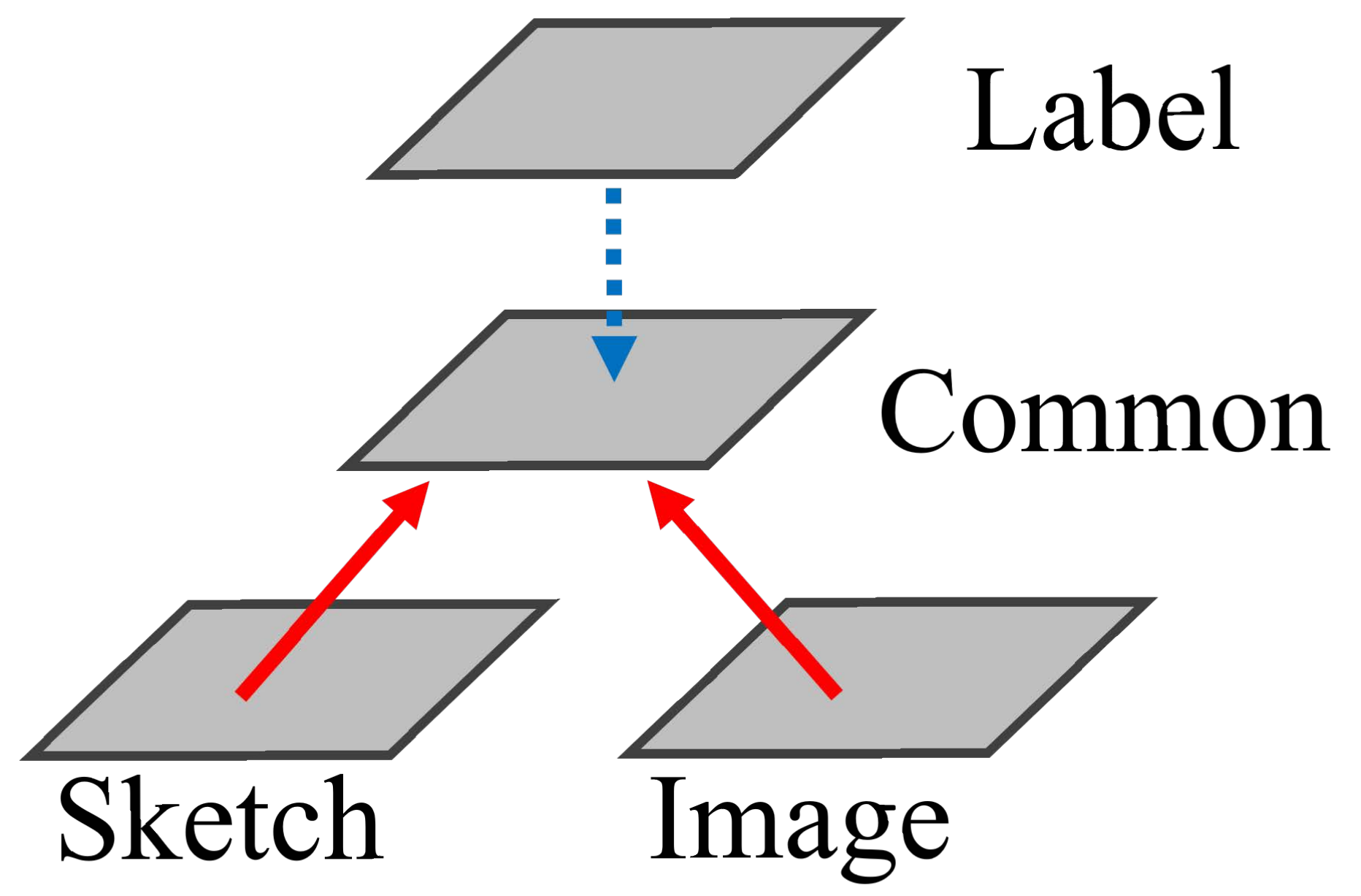}
				\label{1}
			\end{minipage}%
		}%
		\subfigure[]{
			\begin{minipage}[t]{2.7cm}
				\centering
				\includegraphics[width=2.7cm]{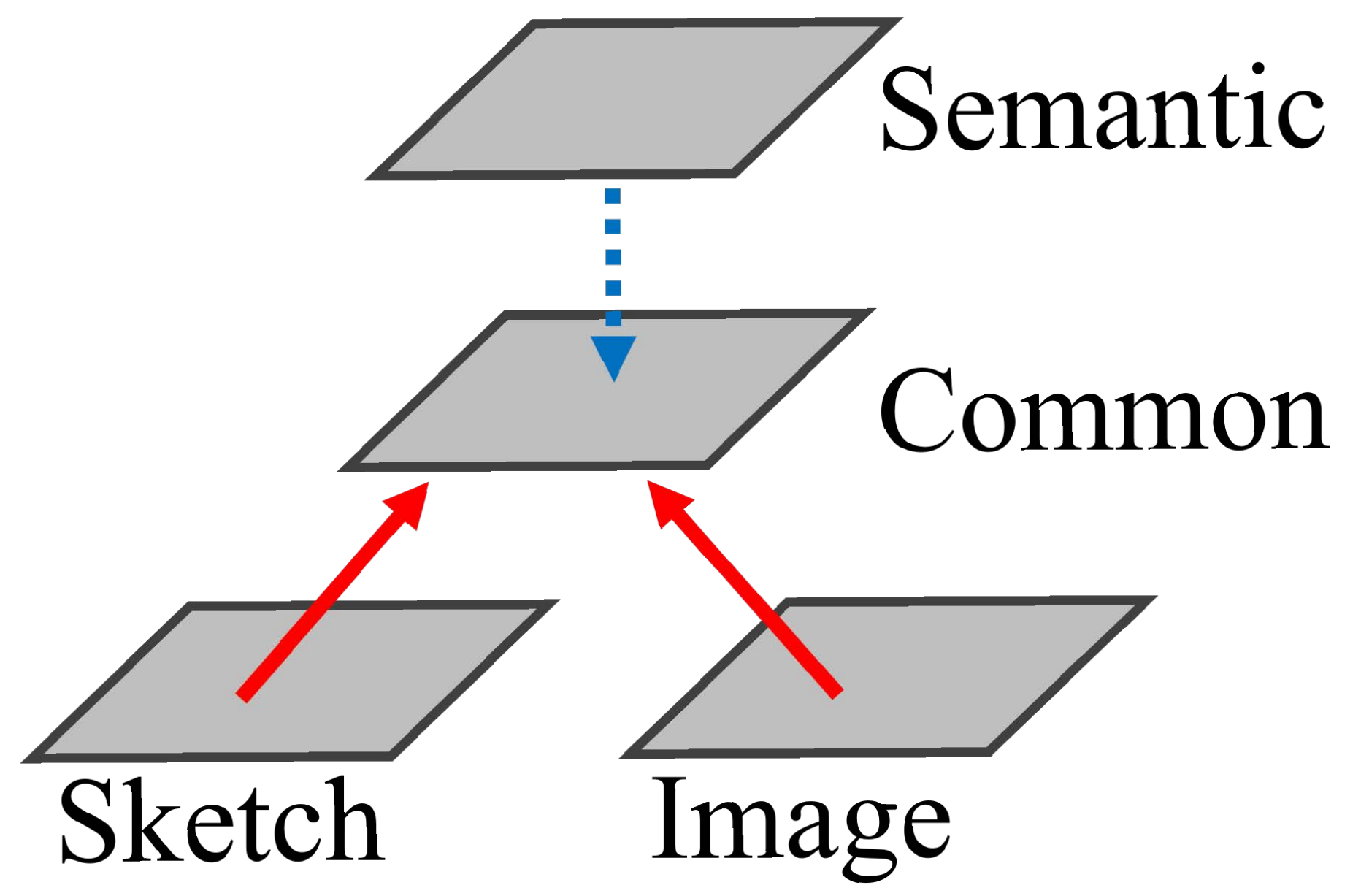}
				\label{2}
			\end{minipage}%
		}%
		\subfigure[]{
			\begin{minipage}[t]{2.7cm}
				\centering
				\includegraphics[width=2.7cm]{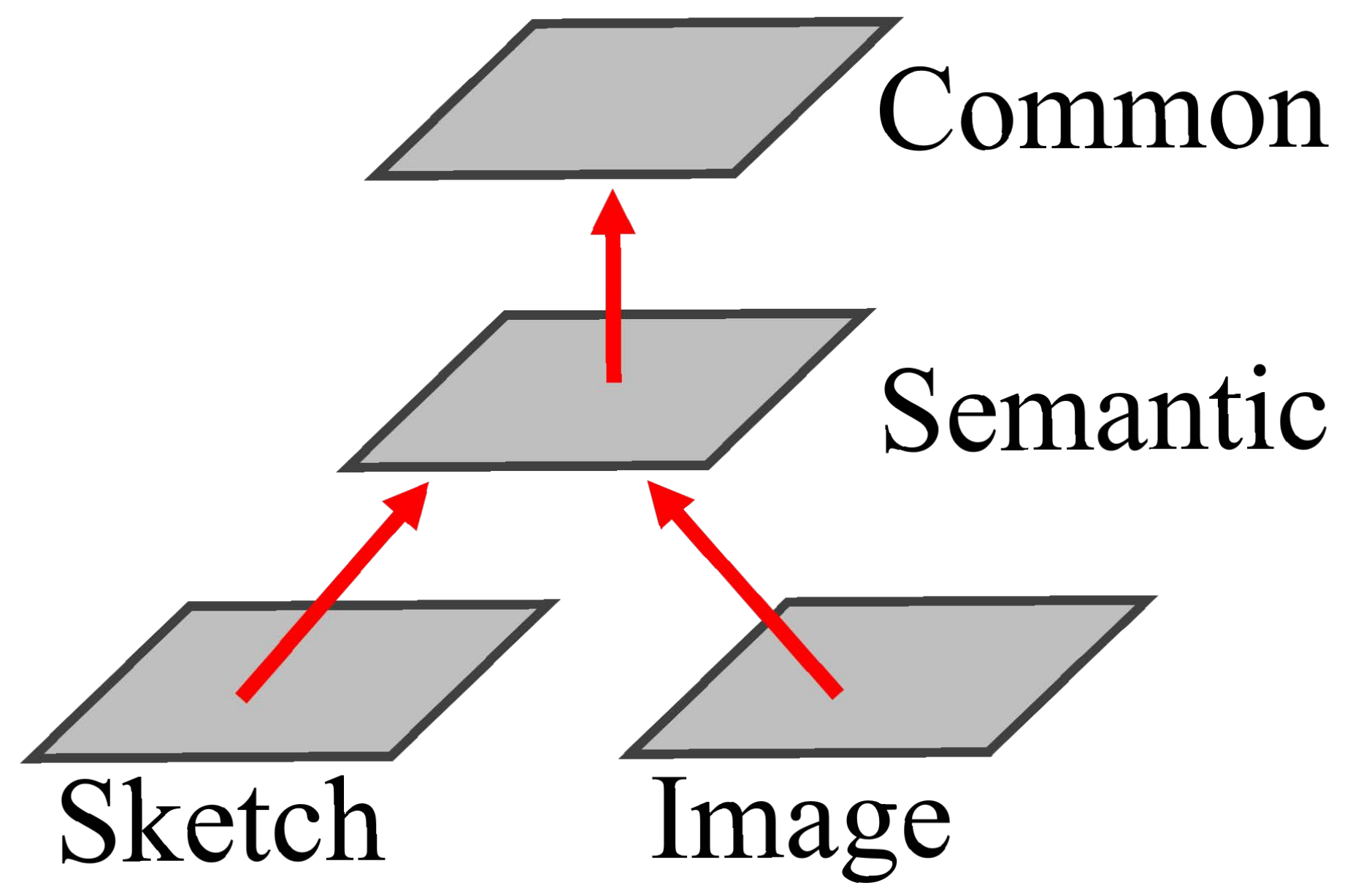}
				\label{3}
			\end{minipage}%
		}%
		
		\centering
		\caption{The ways in Figure~\ref{1} and ~\ref{2} simultaneously project visual features and label/semantic supervision into a low-dimensional common space for efficient retrieval. The Figure 1(c) shows our way that first aligns sketch and image to semantic embedding explicitly and then project them into common space.}
		\label{introduction}
		
	\end{figure}
	
	\begin{figure*}[!ht]
		\centering
		\includegraphics[width=16.5cm]{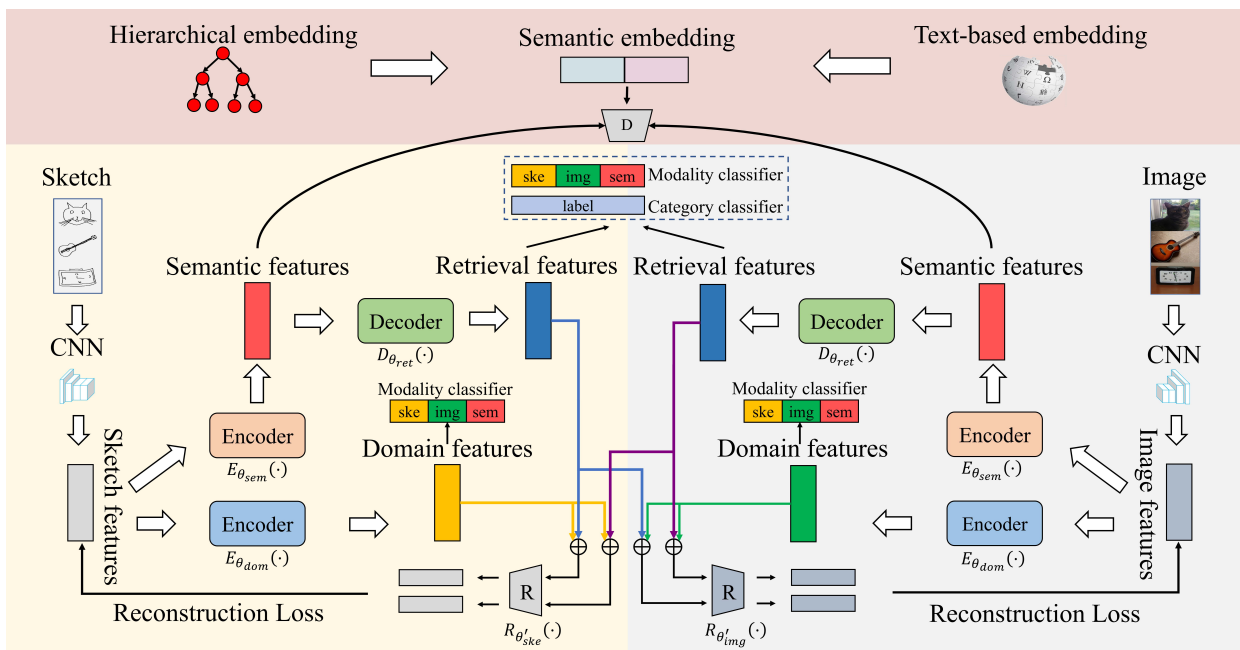}
		\caption{Flowchart of the proposed progressive domain-semantic feature decomposition network. First, it decomposes the visual features into semantic features and domain features, where the semantic features are learned in an adversarial fashion, while the domain features are learned under the constraint of modality classifier. Subsequently, the semantic features are projected into a common space as retrieval features. Moreover, the semantic embedding derived from the text-based embedding and hierarchical embedding serve as true examples to the discriminator. Meanwhile, the cross-reconstruction loss guarantees that the retrieval features only contain high-level knowledge, which is beneficial to reducing the interference of domain features.}
		\label{fig_framework}
	\end{figure*}
	As shown in Figure~\ref{1} and \ref{2}, previous works~\cite{dutta:semantically} attempted to overcome these challenges through simultaneously projecting sketch/image features and label/semantic supervision to a low-dimensional common space.
	However, these strategies deteriorates the original semantic knowledge, since the low-dimensional projection maps complete semantic embedding from original semantic space to semantically incomplete low-dimensional space, causing PDFD unable to transfer knowledge well when learning semantic from two modalities.
	Hence, as shown in Figure~\ref{3}, we present our progressive projection strategy that first learns semantic features with original semantic supervision, and then projects them into a common retrieval space, which is beneficial to knowledge transfer due to the strong semantic supervision can be maintained.
	Another issue is that, the domain information and semantic information are entangled in visual features, and the distribution of two domains are highly heterogeneous, which seriously hinders the reduction of domain gap between sketch and image, making cross-modal matching difficult.
	Since semantic information remains the same expression in different domains, we argue that only semantic information is crucial for cross-modal matching.
	To this end, we decompose the visual features to attain domain-independent retrieval features which only contain clean and complete semantic information.

	In this paper, we have proposed a \underline{P}rogressive \underline{D}omain-independent \underline{F}eature \underline{D}ecomposition (PDFD) network for ZS-SBIR task.
	First, PDFD decomposes visual features into semantic features and domain features, where the semantic features are adversarially learned with the supervision of original semantic embedding, while the domain features are learned with a modality classifier.
	Subsequently, the learned semantic features are projected into a common space as retrieval features under the category and modality supervision.
	Besides, in order to reduce the domain interference for cross-modal matching, we introduce cross-reconstruction loss to encourage the retrieval features capture clean and complete semantic information.
	It is expected that such retrieval features can reconstruct the sketch or image visual features by combining with sketch-domain features or image-domain features.
	In the network, the parameters of decoders and encoders for sketches and images are shared.
	
	The main contributions of this work are summarized:
	\begin{itemize}
		\item We propose a feature decomposition model to effectively reduce the domain gap by generating domain-independent retrieval features with a novel cross-reconstruction loss.
		\item The proposed progressive projection strategy preserves the strong semantic supervision when generating retrieval features, which is beneficial to knowledge transfer under the zero-shot scenario.
		\item Extensive experiments conducted on two popular large-scale datasets demonstrate that our proposed PDFD significantly outperforms state-of-the-art methods.
	\end{itemize}

	\section{Related Work}
	In this section, we briefly review the prior literature in the fields of SBIR, ZSL and ZS-SBIR.
	\subsection{Sketch-Based Image Retrieval}
	The existing SBIR approaches can be mainly divided into two categories: hand-crafted features based methods and deep learning based ones. The hand-crafted features based methods attempt to bridge the domain gap by using edge-maps extracted from images, such as gradient field HOG descriptor~\cite{hu:performance} and Learned Key Shapes (LKS)~\cite{saavedra:sketch}. As for the deep learning based methods, Yu~\emph{et al.}~\shortcite{yu:sketch} first adopted CNN to learn better feature representation for sketches and images. Besides, siamese architecture~\cite{qi:sketch} achieves a better metric of retrieval by minimizing the loss function for samples from the same category and maximizing the loss function for samples from different categories.
	
	\subsection{Zero-Shot Learning}
	Existing zero-shot approaches can be classified into two categories: embedding-based and generative-based approaches. In the first category, some approaches learn non-linear multi-modal embedding~\cite{akata:label,xian:latent}. As for generative-based approaches, a conditional generative moment matching network~\cite{bucher:generating} adopts generator learning with seen class to generate unseen class , which turns ZSL into supervised learning problems. Moreover, some studies~\cite{xian:latent,akata:evaluation} utilize auxiliary information, \eg a text-based embedding~\cite{mikolov:efficient} or a hierarchical embedding~\cite{miller:wordnet} for label embedding, which is beneficial to knowledge transfer.
	
	\subsection{Zero-Shot Sketch-Based Image Retrieval}
	The first work~\cite{shen:zero} of ZS-SBIR utilizes a multi-modal learning network to mitigate heterogeneity between two different modalities. The recent work SEM-PCYC~\cite{dutta:semantically} proposes a paired cycle-consistent generative model based on semantically alignment, which maintains a cycle consistency that only requires supervision at category level. Besides, SAKE~\cite{liu:semantic} proposes a teacher-student network to maximally preserving previously acquired knowledge to reduce the domain gap between the seen source domain and unseen target domain.
	
	\section{Methodology}
	\subsection{Problem Definition}
	We first provide a formal definition of the ZS-SBIR task. Let $\mathcal{D}_{tr}=\{\mathcal{X}^{seen}, \mathcal{Y}^{seen}, \mathcal{S}^{seen}, \mathcal{C}^{seen}\}$ be a training set that contains sketches $\mathcal{X}^{seen} = \{x_{i}^{seen}\}_{i = 1}^{N_s}$, images $\mathcal{Y}^{seen} = \{y_{i}^{seen}\}_{i = 1}^{N_s}$, semantic embedding $\mathcal{S}^{seen} = \{s_{i}^{seen}\}_{i = 1}^{N_s}$, and category labels $\mathcal{C}^{seen} = \{c_{i}^{seen}\}_{i = 1}^{N_s}$ with $N_s$ samples. The test set is denoted as $\mathcal{D}_{te}=\{\mathcal{X}^{un}, \mathcal{Y}^{un}, \mathcal{C}^{un}\}$ that contains sketches $\mathcal{X}^{un} = \{x_{i}^{un}\}_{i = 1}^{N_u}$, images $\mathcal{Y}^{un} = \{y_{i}^{un}\}_{i = 1}^{N_u}$ and category labels $\mathcal{C}^{un} = \{c_{i}^{un}\}_{i = 1}^{N_u}$ with $N_u$ samples, which satisfies the zero-shot setting $\mathcal{C}^{seen}\cap \mathcal{C}^{un} = \varnothing$. During the test, given an unseen sketch $x_{i}^{un}$ in $\mathcal{X}^{un}$, the objective of ZS-SBIR is to retrieve corresponding natural images from the test image gallery $\mathcal{Y}^{un}$.
	
	The architecture of our proposed PDFD is illustrated in Figure~\ref{fig_framework}, which contains two branches for sketches and images, respectively. Each branch first decomposes visual features into domain features and semantic features. Subsequently, decoders with shared parameters are trained to project semantic features into retrieval features for subsequent ZS-SBIR task.

	\subsection{Visual Features Decomposition}
	In zero-shot learning, it is important to provide knowledge supervision when learning semantic features. Our proposed PDFD utilizes text-based embedding and hierarchical embedding to provide such supervision.
	
	\subsubsection{Semantic Knowledge Embedding}
	In PDFD, we adopt two widely-used text-based embedding, \ie Word2Vec~\cite{mikolov:efficient} and GloVe~\cite{pennington:glove} to obtain text representations.
	As for hierarchical embedding in PDFD, the similarity between words is calculated in WordNet\footnote{https://wordnet.princeton.edu/} with path similarity and Jiang-Conrath~\cite{jiang:semantic} similarity.
	
	\subsubsection{Semantic Features}
	As illustrated in Figure~\ref{fig_framework}, each branch has a semantic encoder $E_{\theta_{sem}}$, common discriminator $D_{\theta_{dis}}$, and semantic embedding that combine text-based embedding and hierarchical embedding. Given a training sketch-image pair, their visual features are extracted from VGG16~\cite{simonyan:very} network pre-trained on ImageNet~\cite{deng:imagenet} dataset (before the last pooling layer). Then the semantic features are learned in an adversarial fashion, which means that the learned semantic features are expected to be as similar as the semantic embedding by `fooling' the discriminator $D_{\theta_{dis}}$. Specifically, the objective can be formulated as:
	\begin{equation}
	\begin{split}
	\mathcal{L}_{adv} = & 2\times\mathbb{E}_{s^{seen}}[\log D_{\theta_{dis}}(s^{seen})]\\
	& \;\,+\mathbb{E}_{x^{ske}}[\log(1-D_{\theta_{dis}}(E_{\theta_{sem}}(x^{ske})))]\\
	& \;\,+\mathbb{E}_{x^{img}}[\log(1-D_{\theta_{dis}}(E_{\theta_{sem}}(x^{img})))],
	\end{split}
	\label{eq-adv}
	\end{equation}
	where $x^{ske}$, $x^{img}$, $s^{seen}$, $E_{\theta_{sem}}(\cdot)$ and $D_{\theta_{dis}}(\cdot)$ denote the sketch features, image features, semantic embedding, semantic generation function, and discriminator function, respectively. Besides, the semantic generation function $E_{\theta_{sem}}(\cdot)$ and discriminator function $D_{\theta_{dis}}(\cdot)$ are parameterized by $\theta_{sem}$ and $\theta_{dis}$. Here, $E_{\theta_{sem}}(\cdot)$ minimize the objective against an opponent $D_{\theta_{dis}}(\cdot)$ that tries to maximize it.
	
	\subsubsection{Domain Features}
	Since semantic features and domain features are separated, we argue that semantic, image-domain features and sketch-domain features should also be distinguished from each other.
	Thus, we categorize these three kinds of features into three different modalities.
	Here, the domain encoder $E_{\theta_{dom}}$ is adopted to attain the domain features with the constraint of the modality classifier. The modality classification loss can be formulated as:
	\begin{equation}
	\begin{split}
	\mathcal{L}_{dmcls}= &-\mathbb{E}[\log P(y^{ske}|x^{dom}_{ske})]\\
	&-\mathbb{E}[\log P(y^{img}|x^{dom}_{img})],
	\end{split}
	\label{eq-dcls}
	\end{equation}
	where $y^{ske}$ and $y^{img}$ are labels indicating whether the corresponding features belong to sketch and image. Moreover, ${x}^{dom}_{ske}$, and $x^{dom}_{img}$ denote the domain features from sketch and image branch respectively. They can be formulated as:
	\begin{equation}
	{x}^{dom}_{ske} = {E}_{\theta_{dom}}(x^{ske}),
	\label{eq-3}
	\end{equation}
	\begin{equation}
	{x}^{dom}_{img} = {E}_{\theta_{dom}}(x^{img}).
	\label{eq-4}
	\end{equation}
	
	It is worth noting that the domain features have also been constrained to cross-reconstruction loss, which will be introduced in Section~\ref{sec:11}.

	\subsection{Retrieval Features Generation}
	\label{sec:11}
	After learning semantic features and domain features, PDFD generates retrieval features under two kinds of constraints.
	\subsubsection{Classification Constraint}
	It should be noted that the semantic features learned from two branches are only constrained by adversarial loss, which only ensures that the semantic features possess semantic knowledge. However, it can not guarantee the features to be class-discriminative. Therefore, category classifier is introduced after the two branches. The category classification loss can be formulated as:
	\begin{equation}
	\begin{split}
	\mathcal{L}_{ccls} = -\mathbb{E}[\log P(y|x^{ret}_{ske})]-\mathbb{E}[\log P(y|x^{ret}_{img})],
	\end{split}
	\label{eq-ccls}
	\end{equation}
	where $y$ is the category label of $x^{ske}$ and $x^{img}$. Moreover, ${x}^{sem}_{ske}$, ${x}^{sem}_{img}$, $x^{ret}_{ske}$ and $x^{ret}_{img}$ denote the semantic features and retrieval features generated from sketch and image branch respectively. The generation of these two features can be formulated as:
	\begin{equation}
	{x}^{ret}_{ske} = {D}_{\theta_{ret}}(x^{sem}_{ske}),
	\label{eq-3}
	\end{equation}
	\begin{equation}
	{x}^{ret}_{img} = {D}_{\theta_{ret}}(x^{sem}_{img}),
	\label{eq-4}
	\end{equation}
	where ${D}_{\theta_{ret}}(\cdot)$ is the generation function of retrieval features.
	
	On the other hand, retrieval features should be domain-independent, such that they ought to be classified the semantic modality. We adopt the same modality classifier as above to ensure that, where the modality classification loss is written as:
	\begin{equation}
	\begin{split}
	\mathcal{L}_{rmcls}= &-\mathbb{E}[\log P(y^{sem}|x^{ret}_{ske})]\\
	&-\mathbb{E}[\log P(y^{sem}|x^{ret}_{img})],
	\end{split}
	\label{eq-dcls}
	\end{equation}
	where $y^{sem}$ is label indicating whether the corresponding features belong to semantic modality.
	\begin{table*}[!t]
		\renewcommand\arraystretch{1.3}
		\centering
		\caption{ZS-SBIR performance of our proposed PDFD compared with existing SBIR, ZSL and ZS-SBIR approaches.}
		\label{compare:result}
		\begin{tabular}{|p{1.2cm}<{\centering}|p{6.5cm}<{\centering}|p{1.3cm}<{\centering}|p{1.3cm}<{\centering}|p{1.3cm}<{\centering}|p{1.3cm}<{\centering}|p{1.3cm}<{\centering}|}
			\hline
			\multirow{2}*{}&\multicolumn{1}{p{6.5cm}<{\centering}|}{\multirow{2}*{Methods}}&\multicolumn{1}{p{1.5cm}<{\centering}|}{\multirow{1}*{Feature}}&\multicolumn{2}{c|}{Sketchy}&\multicolumn{2}{c|}{TU-Berlin}\\\cline{4-7}
			&&\multicolumn{1}{p{1.5cm}<{\centering}|}{\multirow{1}*{Dimension}}&{mAP@all}&{Prec@100}&{mAP@all}&{Prec@100} \\
			\hline
			\multirow{7}*{SBIR}&Siamese CNN~\cite{qi:sketch}&64&0.132&0.175&0.109&0.141\\
			
			&SaN~\cite{yu:sketch}&512&0.115&0.125&0.089&0.108\\
			
			&GN Triplett~\cite{sangkloy:sketchy}&1024&0.204&0.296&0.175&0.253\\
			
			&3D Shape~\cite{wang:community}&64&0.067&0.078&0.054&0.067\\
			
			&DSH (binary)~\cite{liu:deep}&64&0.171&0.231&0.129&0.189\\
			
			&GDH (binary)~\cite{zhang:generative}&64&0.187&0.259&0.135&0.212\\
			\hline
			\multirow{4}*{ZSL}&DeViSE~\cite{frome:devise}&300&0.067&0.077&0.059&0.071\\
			
			&JLSE~\cite{zhang:zero}&100&0.131&0.185&0.109&0.155\\
			
			&SAE~\cite{kodirov:semantic}&100&0.216&0.293&0.167&0.221\\
			
			&ZSH (binary)~\cite{yang:zero}&64&0.159&0.214&0.141&0.171\\
			\hline
			\multirow{10}*{ZS-SBIR}&ZSIH (binary)~\cite{shen:zero}&64&0.258&0.342&0.223&0.294\\
			
			&CVAE~\cite{kiran:zero}&4096&0.196&0.284&0.005&0.001\\
			
			&SEM-PCYC~\cite{dutta:semantically}&64&0.349&0.463&0.297&0.426\\
			
			&SEM-PCYC(binary)~\cite{dutta:semantically}&64&0.344&0.399&0.293&0.392\\
			
			&SAKE(binary)~\cite{liu:semantic}&64&0.364&0.487&0.359&0.481\\
			
			&CSDB~\cite{dutta:style}&64&0.484&0.375&0.355&0.254\\
			
			&PDFD (ours)&64&\textbf{0.623}&\textbf{0.726}&\textbf{0.460}&\textbf{0.595}\\
			
			&PDFD (ours binary)&64&\textbf{0.638}&\textbf{0.755}&\textbf{0.386}&\textbf{0.542}\\
			
			&SAKE~\cite{liu:semantic}&512&0.547&0.692&0.475&0.599\\
			
			&PDFD (ours)&512&\textbf{0.661}&\textbf{0.781}&\textbf{0.483}&\textbf{0.600}\\
			
			\hline
		\end{tabular}
		
	\end{table*}

	\subsubsection{Cross Reconstruction Constraint}
	To ensure learning clean semantic-rich retrieval features, we argue that such features should reconstruct the original sketch/image visual features by combined with sketch/image-domain features. To this end, the cross-reconstruction loss is introduced to ensure the reconstructed features are similar to the original features. These reconstructed features are formulated as
	
	\begin{equation}
	\tilde{x}^{ske}_{1} =R_{{\theta_{ske}^{'}}}(x^{ret}_{ske} + x^{dom}_{ske}),
	\label{eq-6}
	\end{equation}
	\begin{equation}
	\tilde{x}^{ske}_{2} =R_{{\theta_{ske}^{'}}}(x^{ret}_{img} + x^{dom}_{ske}),
	\label{eq-7}
	\end{equation}
	\begin{equation}
	\tilde{x}^{img}_{1} =R_{{\theta_{img}^{'}}}(x^{ret}_{ske} + x^{dom}_{img}),
	\label{eq-8}
	\end{equation}
	\begin{equation}
	\tilde{x}^{img}_{2} =R_{{\theta_{img}^{'}}}(x^{ret}_{img} + x^{dom}_{img}),
	\label{eq-9}
	\end{equation}
	where $R_{{\theta_{ske}^{'}}}(\cdot)$ and $R_{{\theta_{img}^{'}}}(\cdot)$ denote the reconstruction function on the sketch branch and image branch, respectively. Besides, $\tilde{x}^{ske}_{1}$ and $\tilde{x}^{ske}_{2}$ denote the reconstructed sketch features; $\tilde{x}^{img}_{1}$ and $\tilde{x}^{img}_{2}$ denote the reconstructed image features. Furthermore, the cross-reconstruction losses in sketch and image branch are written as
	
	\begin{align}
	\label{eq-ske} \mathcal{L}_{rec\_ske} &=||\tilde{x}^{ske}_{1} - x^{ske}\>||_{2}^{2} + ||\tilde{x}^{ske}_{2} - x^{ske}\>||_{2}^{2},\\
	\label{eq-img} \mathcal{L}_{rec\_img} &=||\tilde{x}^{img}_{1} \!- x^{img}||_{2}^{2} + ||\tilde{x}^{img}_{2} - x^{img}||_{2}^{2}.
	\end{align}
	The total cross-reconstruction loss can be formulated as
	\begin{equation}
	\mathcal{L}_{rec} =\mathcal{L}_{rec\_ske} + \mathcal{L}_{rec\_img}.
	\label{eq-sem}
	\end{equation}

	\subsection{Objective and Optimization}
	Since the modality loss is constrained both on domain features and semantic features, we can formulate the total modality loss as:
	\begin{equation}
	\mathcal{L}_{mcls} = \mathcal{L}_{dmcls} + \mathcal{L}_{rmcls}.
	\label{eq-sum}
	\end{equation}
	Finally, the full objective of our proposed PDFD is:
	\begin{equation}
	\begin{split}
	\mathcal{L} =\;&\lambda_{adv}\times\mathcal{L}_{adv}\! + \lambda_{ccls}\,\times\mathcal{L}_{ccls}\\
	+\;&\lambda_{rec}\,\times\mathcal{L}_{rec}+ \lambda_{mcls}\!\times\mathcal{L}_{mcls},
	\end{split}
	\label{eq-sum}
	\end{equation}
	where $\lambda_{adv}$, $\lambda_{ccls}$, $\lambda_{rec}$ and $\lambda_{mcls}$ are coefficients for balancing the overall performance.
	
	\section{Experiment}
	\subsection{Datasets and Setup}
	There are two widely-used large-scale sketch datasets Sketchy~\cite{sangkloy:sketchy} and TU-Berlin~\cite{eitz:humans} for ZS-SBIR.
	
	\textit{\textbf{Sketchy}} originally consists of 75,479 sketches and 12,500 images from 125 categories. Liu \emph{et al.}~\shortcite{liu:deep} extended the image gallery by collecting extra 60,502 images from ImageNet~\cite{deng:imagenet} dataset, such that the total number of images is 73,002 in the extended version.
	
	\textit{\textbf{TU-Berlin}} originally consists of 20,000 unique free-hand sketches evenly distributed over 250 object categories. Compared to Sketchy, TU-Berlin only has category-level matches rather than instance-level matches.
	
	Following the same zero-shot data partitioning in SEM-PCYC~\cite{dutta:semantically}, we also follow the same evaluation criterion in most previous works~\cite{dutta:semantically,shen:zero} in terms of mean average precision (mAP@all) and precision considering the top 100 (Prec@100) retrievals.

	\subsection{Implementation Details}
	Our model is trained with Adam~\cite{kingma:adam} optimizer on PyTorch with an initial learning rate $ = 0.0001$, $\beta_{1}=0.5$, $\beta_{2}=0.99$. The input size of the image is 224$\times$224. The coefficients of each loss are $\lambda_{adv}=1.0$, $\lambda_{rec}=1.0$, $\lambda_{mcls}=1.0$, $\lambda_{ccls}=0.01$ when training on Sketch and $\lambda_{adv}=1.0$, $\lambda_{rec}=0.5$, $\lambda_{mcls}=0.4$, $\lambda_{ccls}=0.4$ when training on TU-Berlin.
	
	VGG16~\cite{simonyan:very} pre-trained on the ImageNet dataset is adopted as a feature extractor. The word text-based embedding~\cite{mikolov:efficient} is adopted to extract 300-dimensional word vectors. Under the zero-shot setting, we only consider the seen classes when constructing the hierarchy embedding~\cite{miller:wordnet} for obtaining the class embedding. Therefore, the hierarchical embedding for Sketchy and TU-Berlin datasets respectively contain 354 and 664 nodes.

	\begin{figure*}[!tb]
	\centering
	\subfigure[Retrieval results on Sketchy.]{
		\label{Fig.sub.1}
		\includegraphics[width=8.6cm]{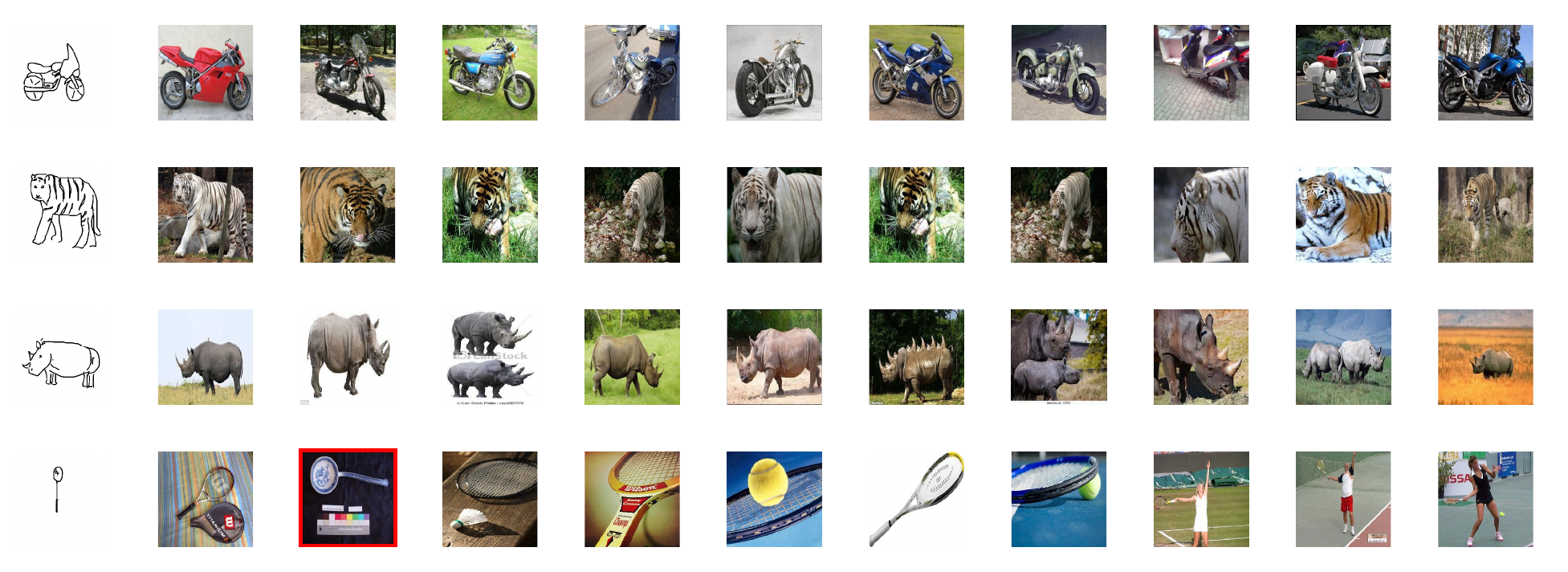}
	}
	\subfigure[Retrieval results on TU-Berlin.]{
		\label{Fig.sub.2}
		\includegraphics[width=8.6cm]{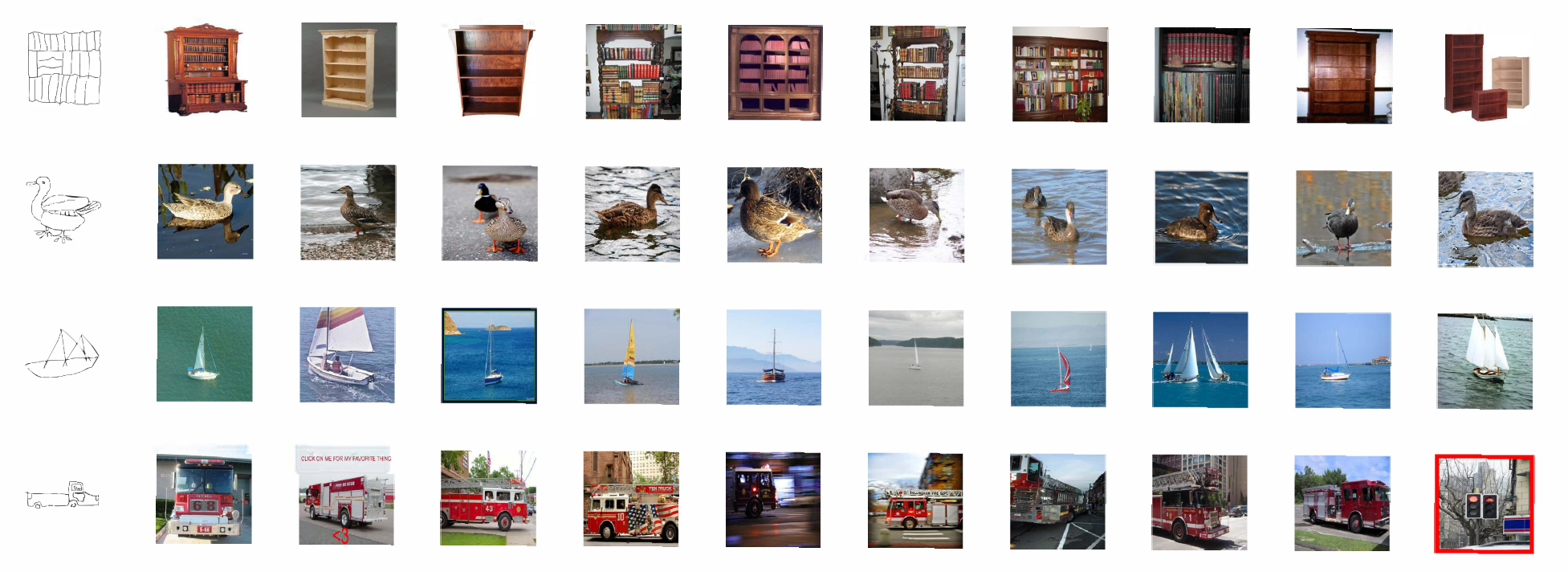}}
	\caption{The top 10 images retrieved by our PDFD on the two datasets. The red borders indicate mis-retrieved images.}
	\label{res:retrieval}

\end{figure*}
\begin{table*}[th]
	\caption{The mAP@all results of PDFD using different semantic embedding and their combinations for ZS-SBIR.}
	\label{coo}
	\begin{center}
		\begin{tabular}{|c|c|c|c|c|c|}
			\hline
			\multicolumn{2}{|c|}{Text-based embedding}&\multicolumn{2}{c|}{Hierarchical embedding}&\multirow{2}*{Sketchy}&\multirow{2}*{TU-Berlin}\\\cline{1-4}
			{Glove}&{Word2Vector}&{Path}&{Ji-Cn~\cite{jiang:semantic}}&&\\
			\hline
			\checkmark&&&&0.583&0.387\\
			&\checkmark&&&0.584&0.388\\
			&&\checkmark&&0.603&0.392\\
			&&&\checkmark&0.603&0.393\\
			\checkmark&&\checkmark&&0.615&0.447\\
			\checkmark&&&\checkmark&0.615&\textbf{0.460}\\
			&\checkmark&\checkmark&&0.622&0.447\\
			&\checkmark&&\checkmark&\textbf{0.623}&0.458\\
			\hline
		\end{tabular}
	\end{center}
\end{table*}

\subsection{Comparison with Peer Methods}
	Apart from ZS-SBIR methods, some existing SBIR and ZSL approaches are also involved in retrieval comparison. The performances of all the comparisons are shown in Table~\ref{compare:result}, where we can observe that most ZS-SBIR methods outperform SBIR and ZSL methods while GN Triplet~\cite{sangkloy:sketchy} and SAE~\cite{kodirov:semantic} reach the best performance in SBIR and ZSL, respectively.
	The main reason is that SBIR and ZSL methods are unable to take both domain gap and knowledge transfer into consideration. Therefore, the ZS-SBIR methods have better performance as they possess both the ability of reducing the domain gap and transferring the semantic knowledge. Due to the larger number of classes in TU-Berlin, all involved methods perform relatively worse on this dataset compared with Sketchy.
	
	\begin{table*}[!th]
		\caption{The mAP@all results of ablation study on our PDFD with several baselines for ZS-SBIR.}
		\begin{center}
			\begin{tabular}{|c|l|c|c|}
				\hline
				\#&\multicolumn{1}{c|}{Description}{}&{Sketchy}&{TU-Berlin}\\
				
				\hline			
				1&Baseline&0.377&0.338\\
				
				2&Baseline + Progressive ($\mathcal{L}_{ccls}$) &0.481&0.374\\
				
				3&Baseline + Progressive ($\mathcal{L}_{ccls}$) + Decomposition ($\mathcal{L}_{mcls}$) &0.510&0.396\\
				
				4&Baseline + Progressive ($\mathcal{L}_{ccls}$)  + Decomposition ($\mathcal{L}_{rec}$)&0.613&0.449\\
				
				5&Baseline + Progressive ($\mathcal{L}_{ccls}$) + Decomposition ($\mathcal{L}_{mcls}$ + $\mathcal{L}_{rec}$)&0.623&0.460\\
				
				\hline
			\end{tabular}
			\label{compare:result1}
		\end{center}
	
	\end{table*}

	Most of the ZS-SBIR methods are conducted to retrieve based on 64-dimensional features, so our model generates 64-dimensional retrieval features for retrieval and outperforms the best competitor~\cite{dutta:style} by more than 13\% on Sketchy and 10\% on TU-Berlin. However, the SAKE~\cite{liu:semantic} adopts the 512-dimensional features for retrieval, and then applies the iterative quantization algorithm~\cite{liu:learning} on the feature vectors to obtain the 64-dimensional binary codes. For a fair comparison, our model also obtains 512-dimensional retrieval features and 64-dimensional binary codes. The result shows that our model significantly outperforms SAKE~\cite{liu:semantic} by around 11\% on Sketchy and 1\% on TU-Berlin when adopting 512-dimensional real-valued retrieval features.
	
	All of these demonstrate the effectiveness of our proposed PDFD for domain gap reduction and semantic knowledge transfer. The retrieved images for sketches using our model are shown in Figure~\ref{res:retrieval}. The red borders indicate the incorrectly retrieved images.

	\subsection{Effect of Semantic Knowledge Embedding}
	Since different types of semantic embedding have different impact on performance, we analyze the effects of different semantic embedding as well as different combinations of them on retrieval performance based on 64-dimensional retrieval features. Table~\ref{coo} shows the quantitative results with different semantic embedding and their combinations. As we can see, the combination of Word2Vec and Jiang-Conrath~\cite{jiang:semantic} hierarchical similarity reaches the highest mAP of 62.3\% on Sketchy, while on TU Berlin dataset, the combination of Glove and Jiang-Conrath reaches the highest mAP of 46.0\%.
	Note that we adopt the same embedding setting as above for all ablation studies.
	
	\subsection{Ablation Study}
	Five ablation studies are conducted to validate the effectiveness of our proposed PDFD as exhibited in Table~\ref{compare:result1}, which are:
	1) A baseline that simultaneously projects visual features and semantic supervision into a low-dimensional common space, in contrast to our proposed progressive projection strategy;
	2) Adding our progressive projection strategy to the baseline that first learns semantic features with the original semantic supervision, and then projects them as retrieval features under the category classification loss $\mathcal{L}_{ccls}$;
	3) Further adding our proposed feature decomposition to attain domain-independent retrieval features under the modality classification loss $\mathcal{L}_{mcls}$;
	4) Replacing the modality classification loss $\mathcal{L}_{mcls}$ in ``3)'' with $\mathcal{L}_{rec}$ to validate the effectiveness of cross-reconstruction;
	5) Full PDFD model.
	Moreover, all the retrieval features in ablation studies are 64-dimensional.

	As shown in Table 3, our full PDFD outperforms all baselines.
	The progressive projection strategy in ``2)'' improves baseline performance by around 10\% on Sketchy and 4\% on TU-Berlin, as this strategy benefits semantic knowledge transfer when learning retrieval features.
	By further decomposing visual features into semantic features and domain features under the modality classification loss $\mathcal{L}_{mcls}$, we can derive domain-independent retrieval features and improve the cross-modal retrieval performance.
	Moreover, the proposed cross-reconstruction loss $\mathcal{L}_{rec}$ encourages learning retrieval features with clean and complete semantic information, which improves the performance by a large margin.
	Finally, with all proposed modules, our full PDFD reaches the highest mAP@all of 62.3\% on Sketchy and mAP@all of 46.0\%.

	\section{Conclusion}	
	We have presented a novel progressive domain-independent feature decomposition network to address the problem of ZS-SBIR more effectively.
	On one hand, a progressive projection strategy is exploited to preserve the semantic information with the strong supervision of original semantic knowledge for learning semantic features. On the other hand, the cross-reconstruction loss is imposed to reduce the domain gap by ensuring that the retrieval features capture clean and complete semantic information. Experiments on two large-scale datasets show that our proposed PDFD significantly outperforms existing state-of-the-art methods in ZS-SBIR task.
	\bibliographystyle{named}
	\bibliography{ijcai20}
	
\end{document}